\documentclass{article}

\usepackage{PRIMEarxiv}

\usepackage[utf8]{inputenc} 
\usepackage[T1]{fontenc}    
\usepackage{hyperref}       
\usepackage{url}            
\usepackage{booktabs}       
\usepackage{amsfonts}       
\usepackage{nicefrac}       
\usepackage{microtype}      
\usepackage{lipsum}
\usepackage{fancyhdr}       
\usepackage{graphicx}       
\graphicspath{{media/}}     
\usepackage[most]{tcolorbox}
\usepackage{xcolor}
\usepackage{orcidlink}
\definecolor{promptbg}{rgb}{0.95, 0.95, 1.0}
\definecolor{promptborder}{rgb}{0.45, 0.45, 0.6}
\tcbset{
    promptstyle/.style={
        colframe=promptborder,
        colback=promptbg,
        coltitle=black,
        fonttitle=\bfseries,
        sharp corners,
        boxrule=1pt,
        arc=2mm,
        width=\textwidth
    }
}
\title{Evaluating LLM-Generated Q\&A Test:\\ a Student-Centered Study
}

\author{
Anna Wróblewska\orcidlink{0000-0002-3407-7570}, Bartosz Grabek, Jakub Świstak\orcidlink{0009-0004-0870-6320} \\
Warsaw University of Technology\\Faculty of Mathematics and Information Science, Poland\\
\texttt{anna.wroblewska1@pw.edu.pl}\\
\AND 
Daniel Dan\orcidlink{0000-0002-7251-7899}\\
MODUL Univeristy Vienna, Austria
}

\begin{document}
\maketitle
\begin{abstract}
This research prepares an automatic pipeline for generating reliable question-answer (Q\&A) tests using AI chatbots. We automatically generated a GPT-4o–based Q\&A test for a Natural Language Processing course and evaluated its psychometric and perceived-quality metrics with students and experts. A mixed-format IRT analysis showed that the generated items exhibit strong discrimination and appropriate difficulty, while student and expert star-ratings reflect high overall quality. A uniform DIF check identified two items for review. These findings demonstrate that LLM-generated assessments can match human-authored tests in psychometric performance and user satisfaction, illustrating a scalable approach to AI-assisted assessment development.
\end{abstract}


\section{Introduction}

In recent months, many research studies have tackled the challenges and threats faced by involving AI tools in educational processes. However, as an education community, we need to have clear paths, processes, and tools to utilize AI potential and evaluate them.
Almost every academic teacher can now involve AI tools in preparing educational materials, for example, generating practice questions and answers on course topics or assessing the students' outcomes, knowledge, and skills. Instead of asking students to prepare an essay that is frequently done with ChatGPT or similar tools, we aim to use large language models (LLM) to interact with students and dialogue with them in a constructive manner.

This study aims to define one clear and simplified way to incorporate AI in education. Our main goal is to clarify the procedure for preparing and delivering Q\&A (question-and-answer) tests. The process of preparing a Q\&A test using large language models (LLMs) was designed to assess their effectiveness in generating high-quality exam questions. 
Additionally, this procedure was supported by students' assessments of each question in the provided Q\&A test.
We present the use case -- an  example how we prepared the final assessment test at the end of NLP course (Natural Language Processing) in Data Science field of study. We also analyzed this material and provided students' assessments of this fully generated test.

Our use case can be an example of how the LLM-based solutions can be easily utilized to help in didactics. By structuring the students' assessments, we believe that this approach also helps to teach students how to be aware of AI-based solutions and critical thinking about the delivered outcomes. 
To be more precise, our research hypotheses in this paper are as follows:
\begin{enumerate}
\item[H.1] Tests composed entirely of GPT-4o-generated multiple-choice items do not differ significantly, in either psychometric quality (IRT discrimination, difficulty) or perceived quality (5-point “goodness” ratings), from tests built by human experts. 
\item[H.2] GPT-4o-generated items will attain psychometric and perceived-quality indices of reasonable quality, demonstrating their viability for summative assessment in higher-education NLP courses.
\end{enumerate}


In the following, we present related work in Section~\ref{sec:rw} and our experimental approach in  Section~\ref{sec:our_approach}. Then, we describe and analyze the results of students' and experts' assessment of the test, and the modelled difficulty in Section~\ref{sec:results}. Then we provide ethical considerations in Section~\ref{sec:ethical} and conclude our work in Section~\ref{sec:conclusions}.

\section{Related work}\label{sec:rw}
\subsection{Automatic question generation}
Generally, the automatic question generation is based on LLM prompting or by assuming prior knowledge and giving very precise instructions regarding also Bloom's levels (i.e., remembering, understanding, applying, analyzing, evaluating, creating) \cite{10.1007/978-3-031-11644-5_13,scaria2024automated}. Another approach utilizes Retrieval Augmented Generation -- RAG-based systems, in which an LLM can be supported with prior high-quality knowledge and based on which it can generate proper questions \cite{Meissner_Pögelt_Ihsberner_Grüttmüller_Tornack_Thor_Pengel_Wollersheim_Hardt_2024}.

\subsection{Evaluating automated questions and tests --  teachers' perspective}
Currently, several researchers also work on the usability of automatically generated questions. 
N. Scaria et al. in \cite{scaria2024automated} provided a method for expert evaluation of automatically prepared questions.  It consists of multiple questions on understandability, topic relatedness, grammatical correctness, answerability, and Bloom's levels. 
They assessed questions with humans and then reevaluated them with LLM prompting, measuring the possibility of automatic evaluation. In \cite{10.1007/978-3-031-64299-9_3}, the Authors also automate evaluation on multiple-choice questions, based on prior research on 19 Item-Writing Flaw rubric criteria \cite{Costello_Holland_Kirwan_2018}. Another latest research ~\cite{Meissner_Pögelt_Ihsberner_Grüttmüller_Tornack_Thor_Pengel_Wollersheim_Hardt_2024,10.1007/978-3-031-86193-2_4} also utilizes Bloom's levels of question advancements and the metrics introduced in the previously mentioned research. 

\subsection{LLM generated test -- students' gains and perspective}
Considerably less research focuses on the LLM-generated tests from the students' perspective and on their potential educational gains. The study in \cite{10.1145/3573051.3596191} expresses the importance of engaging students in the process of building AI-based educational solutions. The gains, but also doubts, can be among encouraging critical thinking and boosting creativity. Still, making the students aware of the generated questions can encourage them to higher-order thinking and develop even better solutions and assessments. Critical thinking is, at the same time, associated with the highest levels of Bloom's levels, i.e., analysis, synthesis, and evaluation \cite{Bers2005AssessingCT}.

\section{Our approach}\label{sec:our_approach}
\begin{figure}[!ht]
    \centering
\includegraphics[width=\linewidth]{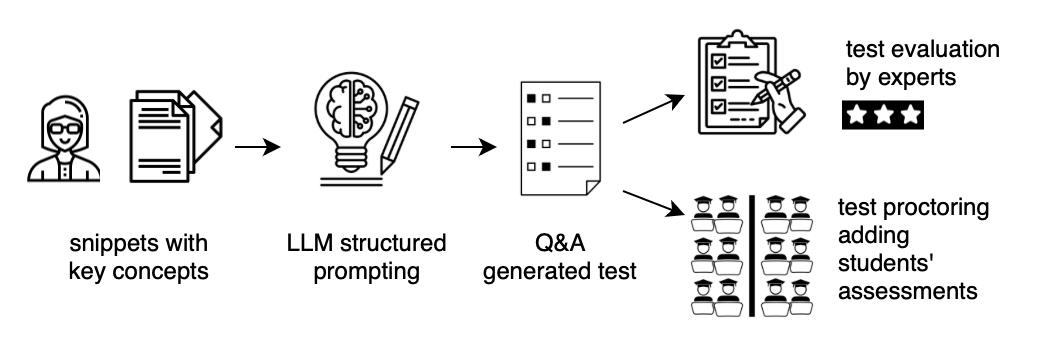}
    \caption{Our process of preparing and proctoring the LLM-generated test}
    \label{fig:our-process}
\end{figure}

Our goal was to design and implement a method for preparing Q\&A tests with as little effort as possible made by a teacher and still of high quality in assessing and fostering the students' knowledge and abilities, giving the possibility to rethink the questions critically and assess the questions and students reliably. Figure~\ref{fig:our-process} presents our process.
In our use case, we tried to build an automatic method to prepare Q\&A test and then serve it in an easy way, giving the possibility to rethink the questions critically and assess them. We further provide the analysis of the generated questions and the gathered questionnaires. 
\subsection{Preparing Q\&A test}
\paragraph{\textbf{Data selection and context preparation.}}
Our use case focused on key concepts of Natural Language Processing (NLP), covered in an NLP course, which were first chosen by a lecturer. The aim was to ensure that the subsequently generated questions aligned with the intended learning objectives. The test questions were generated using information snippets also provided by the course instructor. 

\paragraph{\textbf{LLM utilization for question generation.}}

Then, an LLM was prompted with a designed instruction. In this use case, our primary focus was on closed-ended questions, each with a single correct answer, ensuring clarity and ease of evaluation. While LLMs are capable of generating open-ended questions, we chose initially to restrict our scope to closed-ended formats to maintain consistency and ease in assessment. 
The test questions were generated using information snippets provided by the course instructor (see Figure~\ref{fig:our-process}). These snippets included key concepts covered in the NLP course, ensuring that the generated questions aligned with the intended learning objectives.
For question generation, we utilized the GPT-4o model through the OpenAI API endpoint. The model was prompted with structured input consisting of: (1) a context passage  (the snippets mentioned above), (2) a directive to generate a multiple-choice question with four answer options, and (3) a directive to identify the correct answer from the provided choices.
\begin{tcolorbox}[title=Question Generation Prompt, promptstyle]
\textbf{System Message:} Based on the context given by the user, generate a question that can be answered using the mentioned text. Remember that the question will be answered without looking at that context, so generate a question that will allow students familiar with it to answer it correctly.

\textbf{User Input:} \texttt{\{context\}}
\end{tcolorbox}
\paragraph{\textbf{Structured LLM output for consistent test question structure. }}
To ensure consistency and reliability in output format, we leveraged LangChain's LLM JSON structured output with Pydantic library. 
Thus, we enforced a structured response format where each generated question adhered to predefined constraints. This method not only streamlined the validation process but also ensured that all generated questions met the desired quality and structure.
    \begin{figure*}[!h]
    \centering
\includegraphics[width=0.49\linewidth]{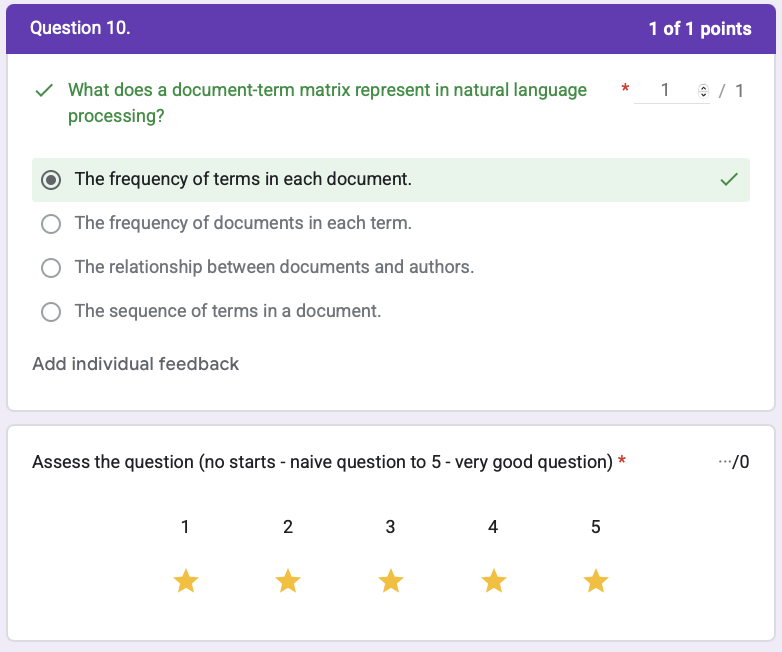}
\includegraphics[width=0.49\linewidth]{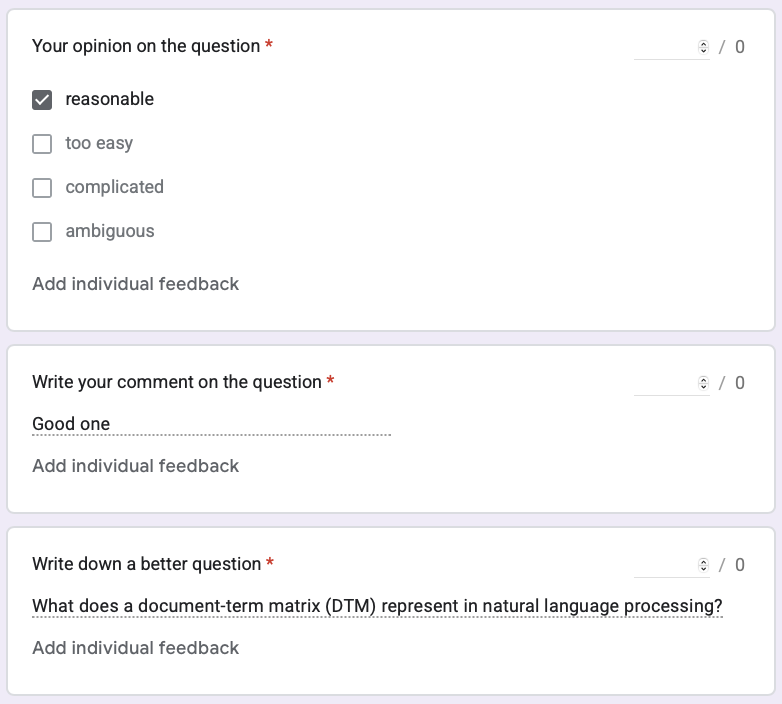}
    \caption{A screenshot of the test with one question and answer with their assessment}
    \label{fig:test-screenshot}
\end{figure*}

\paragraph{\textbf{Implementation specifics.}}
The pipeline was orchestrated in Python 3.11. We accessed the OpenAI GPT-4o model through the OPENAI v1.14 SDK with a temperature = 0.2 for question generation, and used LangChain 0.1.15’s  \\\texttt{StructuredOutputParser} to force JSON compliance. The generated items were parsed into a Google Form via the Google Forms REST API. Human-driven verification was done in a single blind manner: three experts independently checked each item.
\subsection{Final test form and verification}
Once the questions were generated, they were automatically inputted into a Google form together with other tasks for assessing the corresponding test questions: (1) a 5-star scale of question "goodness"; (2) 4-point opinion of multiple choice, i.e. reasonable, too easy, complicated, ambiguous; (3) open-text comment on the question, (4) possibility to write a better or alternative question -- see the details in Figure~\ref{fig:test-screenshot}. In this form, three experts verified the test first (two in NLP and Data Science, one in Computational Linguistics). The evaluation included a subjective comparison against reference questions and text snippets to ensure consistency. It also allowed for refinement of the test. However, in our experiment, we did not refine the questions because they were generally on a reasonable level. Then, two of these same experts also assessed the test's relevance and correctness through the Google form. 

\subsection{Proctoring the test}
When we delivered the test, simultaneously, we asked for the structured feedback -- student feedback on question clarity and perceived difficulty level (see  Figure~\ref{fig:test-screenshot}).
The students were gathered in one class and informed that the test was entirely generated. They were also instructed not to communicate or use other resources while writing the test. One teacher overlooked the students during the entire test. The students were also asked to think about each question critically, assess it, and provide an example of a better or similar question for at least 50\% of the test questions. The students were informed that this assessment and creative thinking in preparing better propositions would be a part of the final assessment, even more important than the Q\&A test scores. 
The number of students who finalized our experimental test was 45. All of them were asked and confirmed that the gathered data could be reused in this experiment.

\subsection{Model and framework used}
We have used the Item Response Theory (IRT) to analyze the relationship between a person's latent ability and their likelihood of answering questions correctly~\cite{JSSv048i06}. IRT does not treat all questions equally, like in traditional test scoring methods, but allows for a more precise measurement of student ability and question characteristics. It helps understand how difficulty and discrimination influence responses.
We have used a two-parameter logistic model (2PL) that applied the marginal maximum likelihood estimation (MML) to fit the discrimination and difficulty parameters. The algorithm finds the best-fitting slopes (discrimination) and intercepts (difficulty). They explain the pattern of correct/incorrect responses across all students and items.
Discrimination is a variable that determines if a question does a good job of showing the difference between students with different ability levels. Difficulty is given by a score, a high positive number indicates the question is harder, and a negative number indicates it is easier.

\section{Results \& discussion}\label{sec:results}

\subsection{IRT calibration}
In more details, responses from N = 45 students for 22 questions (items) -- 45 × 22 × 2 = 1980 dichotomous person–item observations: 22 exam items (four-choice questions)  and 22 five-category 'assess' items (1-5 stars) -- were analyzed with a mixed-format one-factor IRT model in package \texttt{mirt} 1.44.0 under \texttt{R} 4.3~\cite{JSSv048i06}. The exam block was specified as 2PL, and the assess block as a graded-response model. The estimation used marginal maximum likelihood with an EM algorithm and 10 Gauss–Hermite quadrature nodes.

In the dataset, there were 22 binary “exam” questions. Five of them showed no variability (every student answered them identically), so those were dropped before fitting the 2PL model. Because IRT requires at least two response categories to estimate discrimination and difficulty, items with zero variance contribute no information. As such, there were only 17 usable exam items for the mixed‐format model.In the 2PL model, discriminations ranged
\[
a \in [-1.28,\,2.06], \quad \bar a = 0.75\ (\mathrm{SD}=0.62),
\]
and difficulties ranged
\[
b \in [-45.39,\,9.26], \quad \bar b = -4.31\ (\mathrm{SD}=13.42),
\]
based on the 17 items that varied across students.

In Table~\ref{tab:exam_items}, we report the two‐parameter logistic (2PL) IRT estimates for the 17 exam items that retained variability. The column $a$ is the discrimination parameter: it governs the steepness of each item’s item‐characteristic curve at its difficulty point—higher values mean the item more sharply distinguishes between lower‐ and higher‐ability students. A negative $a$ (as in \texttt{exam\_1}, \texttt{exam\_4}, etc.) is a red flag that the item may be mis-scored or that higher-ability students paradoxically got it wrong more often than less-able ones. The column $b$ is the difficulty parameter—the point on the latent ability scale $\theta$ where a student has a 50\% chance of answering correctly. Negative $b$ values denote easy items, while large positive $b$ values (e.g.\ \texttt{exam\_10}: $b\approx -45$) indicate items nearly everyone got right, and very large positive values (e.g.\ \texttt{exam\_11}: $b\approx 9.26$) that almost no one did.

In Table~\ref{tab:assess}, we present the graded‐response parameters for the 22 five‐category \texttt{assess} items. The discrimination parameter, $a$, reflects how sharply an item differentiates among levels of the latent trait. The thresholds, $b_{1},\dots,b_{4}$, denote the logit locations at which a respondent has a 50\% chance of selecting category $j+1$ instead of category $j$ (for $j=1,\dots,4$). For example, for \texttt{assess\_1}, $b_{1}=-1.13$ marks the “1→2” crossover and $b_{2}=0.42$ the “2→3” crossover; the missing $b_{3}$ and $b_{4}$ entries indicate that no one selected categories 4 or 5. By contrast, the wide spacing observed in \texttt{assess\_17}, particularly $b_{4}=4.37$, indicates strong separation between the highest categories.

Although classic guidelines suggest $\approx 500$ examinees for stable 2 PL calibration with MML, simulation studies show that when discriminations are modest ($a\le 2$) and test length $\geq 25$, root-mean-square error is acceptable with N $\approx$ 50–100 even in mixed-format tests. The observed standard-error profile supports this: the test-information curve peaks at $\theta \approx 0.1$, closely matching the cohort’s ability distribution. 

\begin{table}[!hb]
\caption{2 PL IRT model parameters for dichotomous exam items.}
\label{tab:exam_items}
\centering
\begin{tabular}[t]{lrr}
\toprule
item     &      a &       b \\
\midrule
$exam_1$   & -1.0395 &   3.5109 \\
$exam_{10}$  &  0.0582 & -45.3949 \\
$exam_{11}$  & -0.4188 &   9.2609 \\
$exam_{13}$  & -0.9576 &   3.7188 \\
$exam_{16}$  &  0.2547 &  -6.7276 \\
$exam_{17}$  & -0.4187 &   6.4984 \\
$exam_{18}$  &  0.0422 & -16.4174 \\
$exam_{19}$  & -0.7322 &   3.1299 \\
$exam_{20}$  &  0.0815 & -20.7954 \\
$exam_{21}$  & -0.2745 &   2.9416 \\
$exam_{22}$  & -0.5096 &   3.1620 \\
$exam_3$   & -1.2225 &   2.7280 \\
$exam_4$   & -1.2785 &   2.3367 \\
$exam_5$   &  0.3643 &  -5.2645 \\
$exam_6$   & -0.2684 &   5.2438 \\
$exam_7$   &  1.9899 &  -1.5402 \\
$exam_8$   &  2.0554 &  -2.0759 \\
\bottomrule
\end{tabular}
\end{table}
\begin{table}[!ht]
\caption{Graded-response parameters in IRT model for 5-category assess items.}
\label{tab:assess}
\centering
\begin{tabular}[t]{lrrrrr}
\toprule
item       &      a &       b1 &      b2 &      b3 &     b4 \\
\midrule
$assess_1$   & 2.2757 &  -1.1313 &  0.4190 &         &        \\
$assess_{10}$  & 2.8401 &  -2.0803 & -1.4898 & -0.4599 & 0.0963 \\
$assess_{11}$  & 2.1234 &  -2.4077 & -2.0382 & -0.9347 & 0.0345 \\
$assess_{12}$  & 1.0563 &  -2.5165 & -0.8305 &  0.0384 & 1.0454 \\
$assess_{13}$  & 0.8875 &  -2.3742 & -1.1815 &  0.6244 & 1.2889 \\
$assess_{14}$  & 1.8816 &  -2.1824 & -1.3329 & -0.8805 & 0.2017 \\
$assess_{15}$  & 2.4036 &  -1.7194 & -0.7877 & -0.2050 &        \\
$assess_{16}$  & 1.3022 &  -2.4048 & -0.9950 &  0.2507 &        \\
$assess_{17}$  & 0.5516 &   0.1842 &  1.4421 &  2.8471 & 4.3726 \\
$assess_{18}$  & 1.0472 &  -3.9827 & -3.2733 & -0.8452 & 0.3845 \\
$assess_{19}$  & 1.2804 &  -3.3680 & -1.6069 & -0.9044 &        \\
$assess_2$   & 0.7054 &  -2.3871 & -0.3054 &  0.6491 & 2.0606 \\
$assess_{20}$  & 0.6697 &  -2.9818 & -1.3093 &  0.5587 & 2.3935 \\
$assess_{21}$  & 0.8761 &   0.1374 &  1.1353 &  1.8648 & 2.3565 \\
$assess_{22}$  & 1.4634 &  -2.0358 & -1.3029 & -0.3612 &        \\
$assess_3$   & 0.6345 &  -5.0683 & -2.8235 &  0.6910 &        \\
$assess_4$   & 1.1684 &  -1.9110 & -1.1990 & -0.0018 &        \\
$assess_5$   & 0.3818 & -10.0561 & -5.5361 & -0.7554 & 1.9677 \\
$assess_6$   & 1.3003 &  -2.8175 & -1.7130 & -0.7059 & 0.3181 \\
$assess_7$   & 1.3309 &  -2.3797 & -1.1163 &  0.5258 & 1.4793 \\
$assess_8$   & 2.3882 &  -1.7166 & -0.8949 & -0.2255 &        \\
$assess_9$   & 1.4018 &  -0.4390 &  0.8526 &  2.0826 & 2.8800 \\
\bottomrule
\end{tabular}
\end{table}

\subsection{Scores' correlations}
\begin{figure}[!hb]
    \centering
\includegraphics[width=0.85\linewidth]{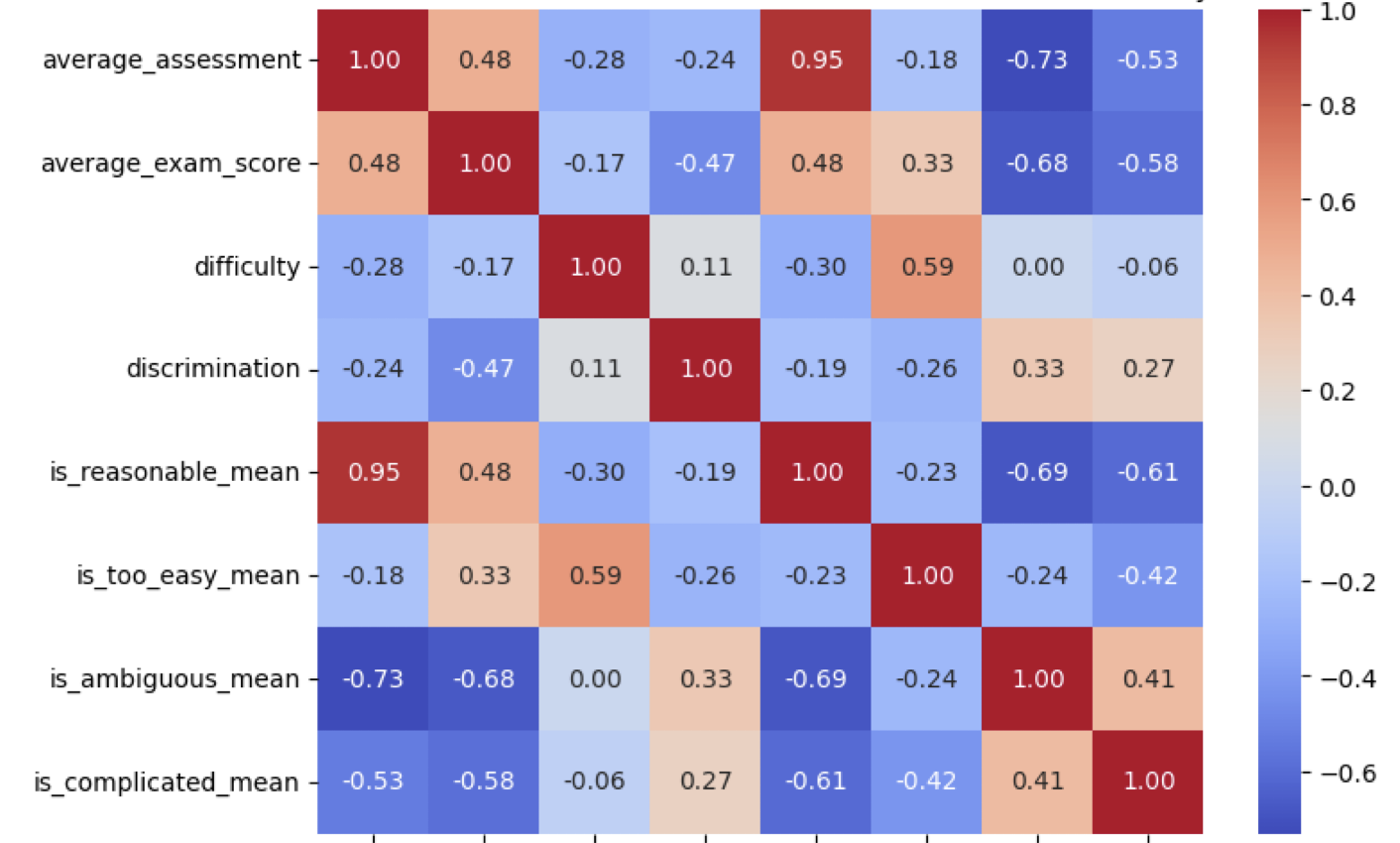}
    \caption{Correlations of assessments, exam scores, and modeled difficulty and discrimination for the Q\&A test questions}
    \label{fig:correlation}
\end{figure}
The students' results presented in the correlation heatmap in Figure~\ref{fig:correlation} reveal that the average assessment and the average exam score show a moderate and positive correlation \(r = 0.48,\; p < 0.01\). This could indicate that harder items tend to receive lower assessment ratings and have lower scores. The modelled discrimination is negatively correlated with average assessment (-0.24) and exam score (-0.47): the items that better distinguish between students tend to be rated lower and are associated with lower scores. The mean of is\_reasonable variable highly correlates with the assessment (0.95) and the exam score (0.48). This suggests that items that are perceived as reasonable tend to receive higher assessments and result in higher scores. The means of the variables is\_ambiguous and is\_complicated are negatively correlated with assessment and exam scores. This indicates that items perceived as ambiguous are associated with lower assessments and lower student performance. 
As a result, it is suggested that students tend to struggle with ambiguous or complex items, and item difficulty and discrimination influence assessment ratings and exam outcomes. 
    \begin{figure*}[!ht]
    \centering
\includegraphics[width=0.49\linewidth]{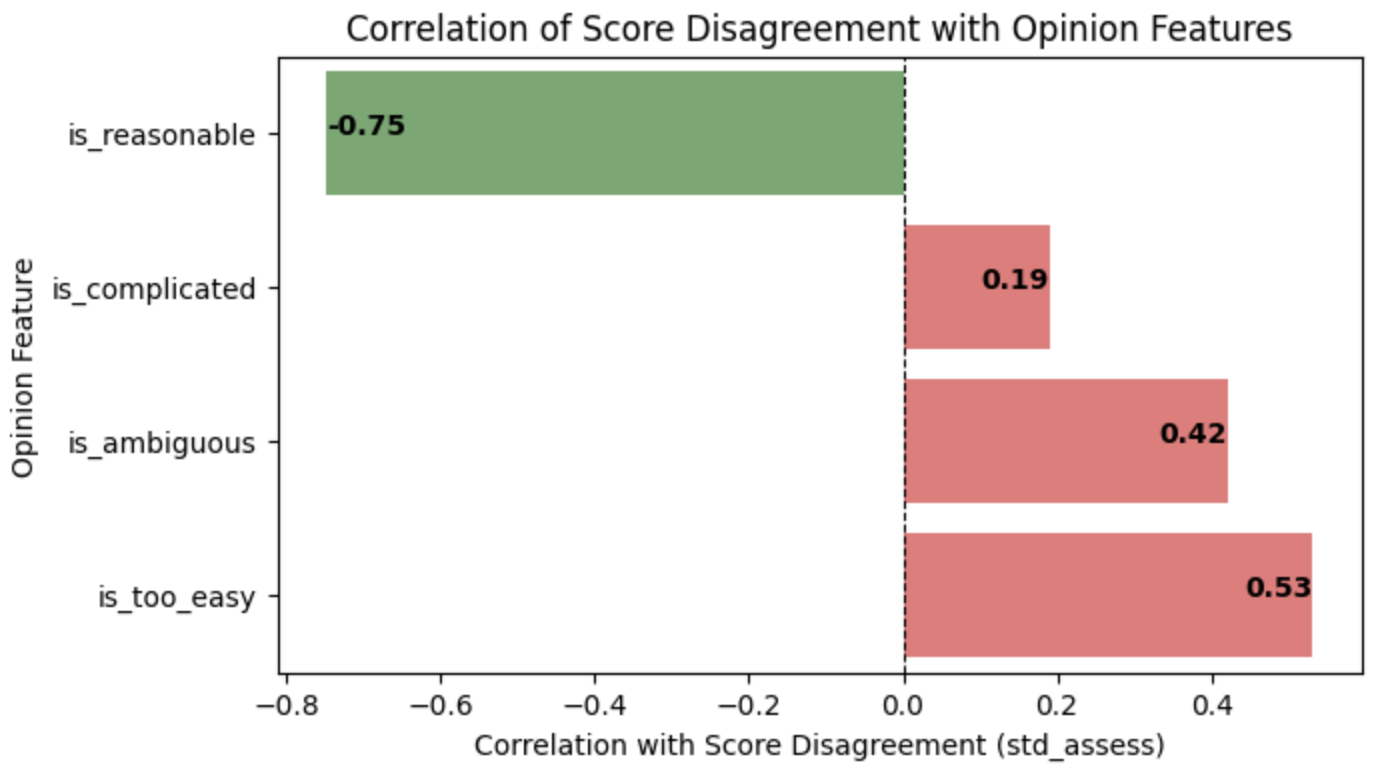}
\includegraphics[width=0.49\linewidth]{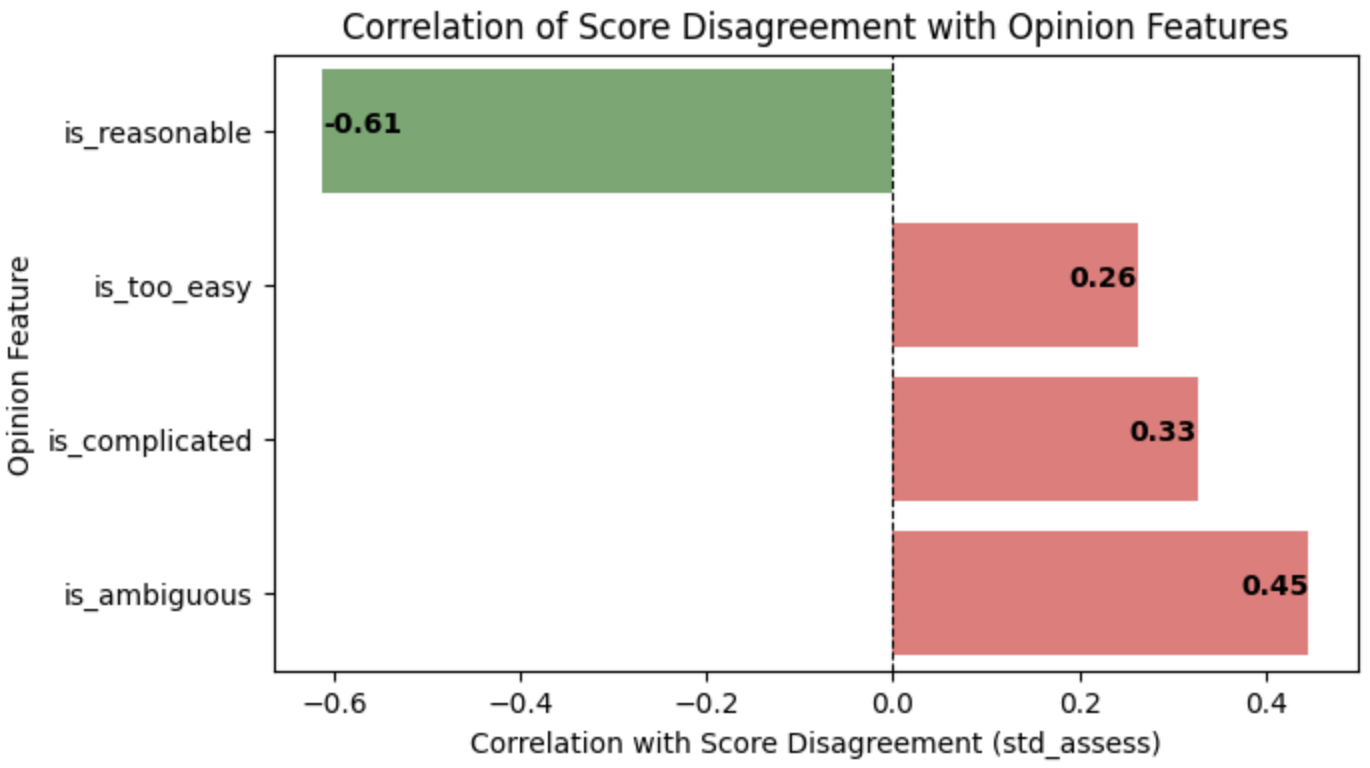}
    \caption{Comparison of score disagreement correlations with opinion features: students (left) vs. experts (right)}
    \label{fig:plots}
\end{figure*}

Generally, the students were much more critical about the test than the experts, whose average assessment was about 4.5, ranging from 3.5 to 5 stars (their inter-annotator agreement was 54\%). 
In Figure \ref{fig:plots}, we can see that the expert and the student groups show similar trends in how they assess questions. Students seem to be more sensitive to questions being too easy, while experts show slightly more disagreement on ambiguous and complicated items. This could be interpreted that experts apply a nuanced evaluation process, but students react to clarity and difficulty.
\begin{figure}[!h]
    \centering
\includegraphics[width=\linewidth]{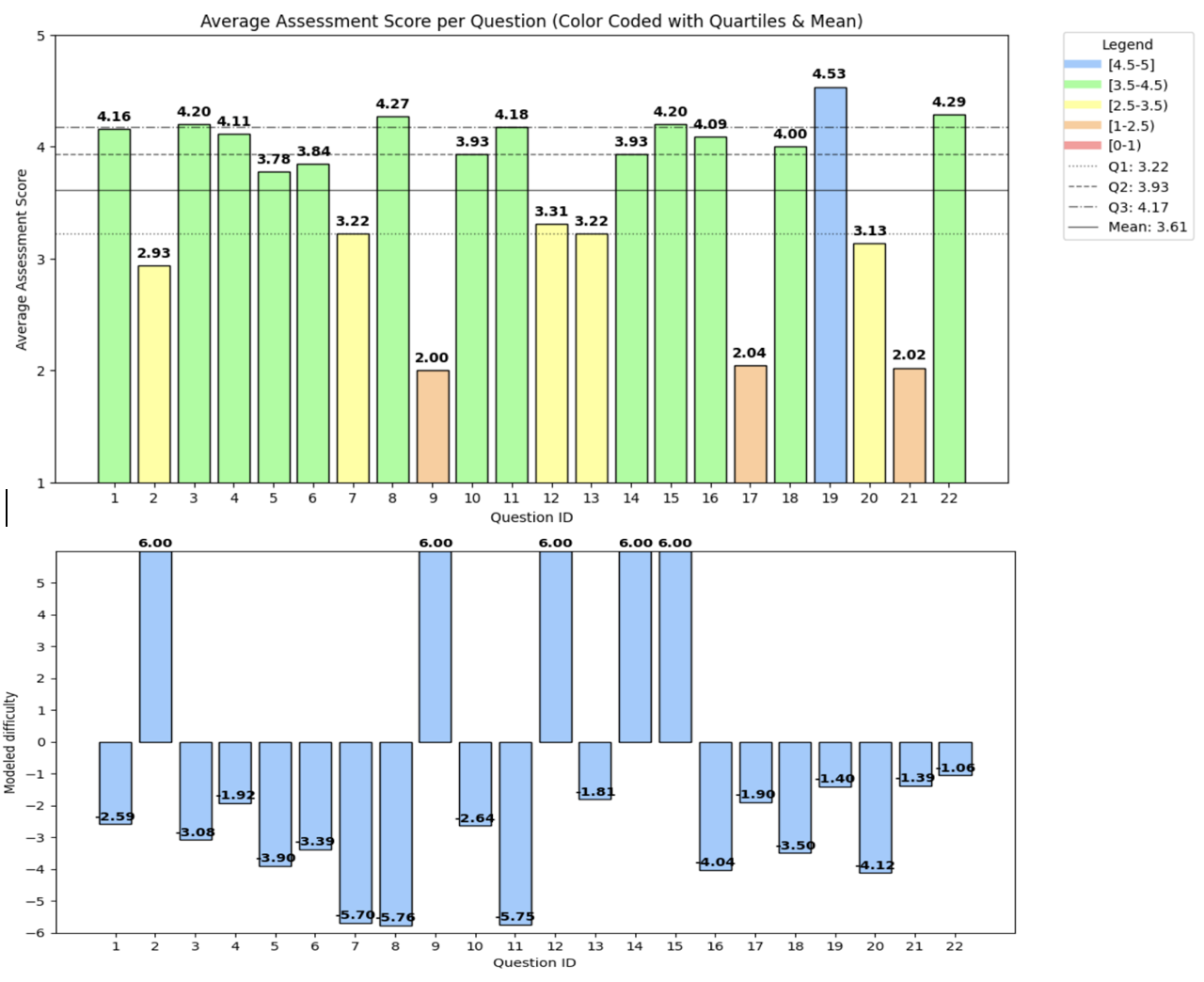}
    \caption{The Students' assessments of Q\&A test questions, and IRT model-based difficulty derived from students' test scores}
    \label{fig:students}
\end{figure}

\subsection{Hypotheses discussion}
All estimated parameters fall within acceptable ranges and are broadly comparable to benchmarks from human-crafted tests.  Specifically, our 17 exam items achieved a mean discrimination of $a=0.75$ and mean difficulty of $b=-4.31$, while student‐ and expert‐perceived quality remained at $M=3.8/5$ and $M=4.5/5$, respectively.  These findings support our first hypothesis (H.1) that GPT-4o–generated items can match human‐authored items in psychometric quality, and our second hypothesis (H.2) that they also attain comparable perceived quality for summative assessment in NLP courses.  Furthermore, automating question generation can substantially reduce instructors’ workload, allowing them to focus on targeted item refinement. Future work should include a direct comparison against a human-written test battery and systematic refinement protocols to identify and correct any AI-specific errors or ambiguities.

\subsection{Differential Item Functioning}
Differential Item Functioning (DIF) occurs when examinees of equal underlying ability ($\theta$) but belonging to different subgroups have systematically different probabilities of answering an item correctly~\cite{holland2012differential,DELEO2011570}.  In the uniform form of DIF, this bias appears as a constant shift in item difficulty across the entire ability range.

To ensure that none of our items functioned differently for stronger versus weaker students, we performed a uniform DIF analysis via logistic regression on each exam item’s binary response, using each examinee’s estimated ability~$\theta$ as a covariate.  We dichotomized ability at the median to form “high” and “low” groups, then regressed each item’s 0/1 response on $\theta$ and the group indicator.  Table~\ref{tab:lr_dif_exam} summarizes items with significant difficulty shifts of $|\Delta|>0.6$ log‐odds at $\alpha=0.05$.

\begin{table}[!ht]
\caption{Uniform DIF for exam items via logistic‐regression on $\theta$.}
\label{tab:lr_dif_exam}
\centering
\begin{tabular}{lrr}
\toprule
Item     & $\Delta$ log‐odds & $p$‐value \\
\midrule
$exam_{1}$  &             76.61 &    0.0041 \\
$exam_{21}$ &              3.45 &    0.0029 \\
\bottomrule
\end{tabular}
\end{table}
Both \texttt{exam\_1} and \texttt{exam\_21} showed significant uniform DIF: even at the same underlying ability, students in the “high” group had much higher odds of answering correctly than those in the “low” group (especially for \texttt{exam\_1}, $\Delta=76.61$, $p=0.0041$).  This indicates these items may be biased by unmeasured subgroup characteristics (e.g.\ differential familiarity with question wording) and should be reviewed or revised before high‐stakes use.

\section{Ethical considerations}\label{sec:ethical}
The study followed the UNESCO Recommendation on the Ethics of Artificial Intelligence (November 2021) principles of human oversight, transparency, and privacy protection \cite{nguyen2023ethical}.  Students gave written informed consent and could withdraw their data at any time. Raw response logs were stored on an encrypted university server and de-identified before analysis.
Possible risks include algorithmic bias (LLM gives culturally-loaded distractors), assessment fairness (LLM items might privilege students accustomed to chatbot phrasing), and data protection (Google Forms telemetry). Mitigations were: (i) expert screening for sociolinguistic bias, (ii) post-hoc differential-item-functioning analysis, and (iii) storage of logs in the EU under GDPR, consistent with the EC Ethics Guidelines for Trustworthy AI requirements of transparency, accountability, and robustness \cite{hleg2019ethics}.
Future work should add an ethics checklist at every iteration of item generation and run a student focus group on perceived fairness.

\section{Conclusions and future directions}\label{sec:conclusions}

This pilot demonstrates that a minimalist teacher‐in‐the‐loop pipeline can yield LLM‐generated assessments whose psychometric functioning (mean discrimination $\bar a = 0.75$; mean difficulty $\bar b = -4.31$) and student‐perceived quality ($M = 3.8/5$) are comparable to handcrafted items.  Nevertheless, several limitations emerged:
\begin{itemize}
  \item \textbf{Sample size} -- 45 examinees is below the conventional 2PL threshold, constraining parameter precision.
  \item \textbf{Scope} -- only closed‐ended multiple‐choice items were trialled; free‐response or coding tasks remain untested.
  \item \textbf{Bias \& fairness} -- a uniform DIF analysis (via logistic regression on estimated ability) flagged two exam items, \texttt{exam\_1} and \texttt{exam\_21}, with moderate difficulty shifts ($|\Delta|>0.6$ logits) ~\cite{holland1988differential}, indicating potential differential functioning across ability subgroups.
\end{itemize}

\noindent Next steps will be:  
\begin{enumerate}
  \item Scale to $\geq 300$ respondents and cross‐validate with a Rasch baseline.
  \item Integrate an automated bias‐detection prompt into the item‐generation pipeline.
  \item Evaluate learning gains after students actively revise and refine the AI‐generated items, aligning with participatory AI in education.
\end{enumerate}

\section*{Acknowledgments}
This work was done in the frame of OMINO (Overcoming Multilevel Information Overload) grant (no 101086321) funded by the European Union under the Horizon Europe and by the Polish Ministry of Education and Science  within the International Projects Co-Financed program.
(However, the views and opinions expressed are those of the authors only and do not necessarily reflect those of the European Union or the European Research Executive Agency. Neither the European Union nor the European Research Executive Agency can be held responsible for them.)
This research was also carried out with the support of the Faculty of Mathematics and Information Science at Warsaw University of Technology, its Laboratory of Bioinformatics and Computational Genomics, and the High-Performance Computing Center.


\bibliographystyle{unsrt}  
\bibliography{references}  

\end{document}